\documentclass[letterpaper, 10 pt, journal, twoside]{IEEEtran}
\IEEEoverridecommandlockouts                              %
\usepackage[style=ieee,doi=false,isbn=false,url=false]{biblatex}
\AtBeginBibliography{\footnotesize}
\usepackage{multicol}
\usepackage[bookmarks=true]{hyperref}

\usepackage{graphicx} %
\usepackage[utf8]{inputenc}
\usepackage{amsmath} %
\usepackage{amssymb}  %
\usepackage{fixmath}  %
\usepackage{array}  %
\usepackage[usenames,dvipsnames]{xcolor}  %
\usepackage{hyperref}
\hypersetup{
    colorlinks,
    linkcolor = BlueViolet,
    urlcolor  = Blue,
    citecolor = Black,
    anchorcolor = Black
}
\usepackage[nohyperlinks,printonlyused,nolist]{acronym}
\usepackage{subcaption}
\usepackage{multirow}

\usepackage{algorithm}
\usepackage{algpseudocode}

\usepackage{tikz}
\usepackage{pgfplots}
\pgfplotsset{compat=newest}
\usepgfplotslibrary{groupplots}

\usepackage{adjustbox}
\usepackage{nicematrix}
\usepackage{setspace}

\begin{acronym}[PCLRHC]
    \acro{MPC}{Model Predictive Control}

    \acro{CVAE}{Conditional Variational Autoencoder}
    \acro{DWA}{Dynamic Window Approach}
    \acro{EB}{Elastic Band}
    \acro{FRP}{Freezing Robot Problem}
    \acro{GP}{Gaussian Process}
    \acrodefplural{GP}{Gaussian Processes}
    \acro{IGP}{Interacting Gaussian processes}
    \acro{IL}{Imitation Learning}
    \acro{IRL}{Inverse Reinforcement Learning}
    \acro{LSTM}{Long-Short Term Memory}
    \acrodefplural{LSTM}{Long-Short Term Memory networks}
    \acro{MAP}{Maximum a Posteriori}
    \acro{MLP}{Multi-Layer Perceptron}
    \acro{MPC}{Model Predictive Control}
    \acro{ORCA}{Online Reciprocal Collision Avoidance}
    \acro{PCLRHC}{Partially Closed Loop Receding Horizon Control}
    \acro{RHC}{Receding Horizon Control}
    \acro{RL}{Reinforcement Learning}
    \acro{ROS}{Robot Operating System}
    \acro{SARL}{Socially Attentive Reinforcement Learning}
    \acro{RGL}{Relational Graph Learning}
    \acro{SOGM}{Spatiotemporal Occupancy Grid Map}
    \acro{TEB}{Timed Elastic Band}
    \acro{VAE}{Variational Autoencoder}
    \acro{CVMM}{Constant Velocity Motion Model}
    \acro{MATS}{Mixtures of Affine Time-varying Systems}

    \acrodef{MID}{Motion Indeterminacy Diffusion}
    \acrodef{T++}{Trajectron++}
    \acrodef{JMID}{Joint Motion Indeterminacy Diffusion}
    \acrodef{iMID}{Individual Motion Indeterminacy Diffusion}
    \acrodef{DDIM}{Denoising Diffusion Implicit Model}
    \acrodef{DDPM}{Denoising Diffusion Probabilistic Model}
    \acrodef{GRU}{Gated Recurrent Unit}
    \acrodefplural{GRU}{Gated Recurrent Units}

    \acro{AV}{Autonomous vehicle}
    \acrodefplural{AVs}{Autonomous vehicles}

    \acrodef{LP}{Linear Program}
    \acrodef{QCQP}{Quadratically Constrained Quadratic Program}
    \acrodef{MPCC}{Mathematical Program with Complementarity Constraints}

    \acrodef{KKT}{Karush-Kuhn-Tucker}
    \acrodef{LICQ}{Linear Independence Constraint Qualification}
    \acrodef{CRCQ}{Constant Rank Constraint Qualification}
    \acrodef{SCQ}{Slater's Constraint Qualification}

    \acrodef{OGM}{Occupancy Grid Map}
    \acrodef{NF}{Navigation Function}
    \acrodef{RVO}{Reciprocal Velocity Obstacle}
    \acrodefplural{RVO}{Reciprocal Velocity Obstacles}
    \acrodef{SICNav}{Safe and Interactive Crowd Navigation}
    \acrodef{SICNav-Diffusion}{Safe and Interactive Crowd Navigation with Diffusion Trajectory Predictions}
    \acrodef{SDICNav}{Safe Diffusion Model Predictive Control for Interactive Crowd Navigation}

    \acrodef{MPC}{Model Predictive Control}
    \acrodef{CAMPC}{Collision Avoiding Model Predictive Control}
    \acrodef{RHC}{Receding Horizon Control}
    \acrodef{CLF}{control-Lyapunov Function}
    \acrodef{DWA}{Dynamic Window Approach}
    \acrodef{SFM}{Social Force Model}
    \acrodef{ESFM}{Extended Social Force Model}

    \acrodef{ORCA}{Optimal Reciprocal Collision Avoidance}
    \acrodef{VO}{Velocity Obstacle}
    \acrodef{CA}{Collision Avoiding}

    \acrodef{ADE}{Average Displacement Error}
    \acrodef{FDE}{Final Displacement Error}
    \acrodef{SADE}{Scene-level Average Displacement Error}
    \acrodef{SFDE}{Scene-level Final Displacement Error}
    \acrodef{KDE}{Kernel Density Estimate}
    \acrodefplural{KDE}{Kernel Density Estimates}
    \acrodef{NLL}{Negative Log Likelihood}

    \acrodef{KF}{Kalman Filter}
    \acrodef{EKF}{Extended Kalman Filter}
    \acrodef{UKF}{Unscented Kalman Filter}
    \acrodef{PF}{Particle Filter}
    \acrodef{SLAM}{Simultaneous Localization and Mapping}

\end{acronym}

\newcommand{\set}[1]{\bigl\{#1\bigr\}}

\newcommand{\norm}[1]{\left\lVert#1\right\rVert}
\newcommand{\pnorm}[2]{\norm{#2}_{#1}}
\newcommand{\twonorm}[1]{\pnorm{2}{#1}}

\newcommand{\Reals}{\mathbb R}

\renewcommand{\v}{\mathbf{v}}
\newcommand{\x}{\mathbf{x}}

\newcommand{\f}{\mathbf{f}}

\newcommand{\Q}{\mathbf{Q}}
\newcommand{\R}{\mathbf{R}}
\renewcommand{\P}{\mathbf{P}}

\newcommand{\trans}[1]{#1^\top}

\usepackage{booktabs}

\newcommand*\centremathcell[1]{\omit\hfil$\displaystyle#1$\hfil\ignorespaces}

\DeclareMathOperator*{\argmin}{arg\,min}
\DeclareMathOperator*{\minimize}{minimize}
\DeclareMathOperator*{\subjectto}{subject\ to}

\newcommand{\bbm}{\begin{bmatrix}}
\newcommand{\ebm}{\end{bmatrix}}
\DeclareMathAlphabet{\mbf}{OT1}{ptm}{b}{n}

\newcommand{\state}{\x}
\newcommand{\humstate}{\tilde{\mathbf{x}}}
\newcommand{\init}{\text{o}}
\newcommand{\stateinit}{{\state_{\init}}}
\newcommand{\sspace}{\mathcal{X}}
\newcommand{\control}{\mathbf{u}}

\newcommand{\cspace}{\Reals^2}
\newcommand{\action}{\control}

\newcommand{\at}[2]{{#1}_{#2}}
\newcommand{\kpone}{{t+1}}
\newcommand{\kmone}{{t-1}}
\newcommand{\stateat}[1]{\at{\state}{#1}}
\newcommand{\robstateat}[1]{\id{r}{\at{\mathbf{x}}{#1}}}
\newcommand{\humstateat}[1]{\at{\humstate}{#1}}
\newcommand{\actionat}[1]{\at{\action}{#1}}
\newcommand{\controlat}[1]{\at{\control}{#1}}

\newcommand{\velvec}{\v}
\newcommand{\vel}{\velvec}

\newcommand{\dist}{{d}}

\newcommand{\robidmarker}{r}
\newcommand{\idRob}{\robidmarker}

\newcommand{\id}[2]{{#2^{(#1)}}}

\newcommand{\pref}{\text{intent}}

\newcommand{\robdyn}{\f}
\newcommand{\humdyn}{\mathbf{h}}
\newcommand{\weightsdyn}{\mathbf{g}}

\newcommand{\numhumans}{N}
\newcommand{\numstatobs}{M}
\newcommand{\forallhumanslongset}{\set{1,\dots,\numhumans}}

\newcommand{\forallhumans}{\forall \idA \in \set{1,\dots,\numhumans}}

\newcommand{\forallstatobsshortset}{\set{\numhumans+1,\dots,\numhumans+\numstatobs}}

\newcommand{\actionhum}{\tilde{\control}}
\newcommand{\actionhumat}[1]{{\at{\actionhum}{#1}}}

\newcommand{\idA}{j}
\newcommand{\idB}{l}
\newcommand{\idStat}{\tilde{l}}

\newcommand{\orcarlxsolnset}{\mathcal{O}}

\newcommand{\horiz}{T}

\newcommand{\stagecostsymb}{l}
\newcommand{\stagecost}[1]{\stagecostsymb(#1)}

\newcommand{\termpenalsymb}{\stagecostsymb_{\horiz}}
\newcommand{\termpenal}[1]{\termpenalsymb(#1)}

\newcommand{\forallactidcs}{\forall t\in\set{0,\dots,\horiz-1}}

\newcommand{\midhisthoriz}{-H}
\newcommand{\numMIDsamples}{S}
\newcommand{\MIDsampleidx}{s}
\newcommand{\predpos}{\MIDsample}
\newcommand{\estpos}{\hat{\MIDsample}}
\newcommand{\MIDsample}{\mathbf{y}}
\newcommand{\MIDSample}{\mathbf{Y}}
\newcommand{\MIDallsamples}{\mathbold{\Upsilon}}
\newcommand{\weight}{w}

\newcommand{\nextBest}[1]{{\underline{#1}}}
\definecolor{blue}{rgb}{0,0,0}
\def\MYTITLE{SICNav-Diffusion: Safe and Interactive Crowd Navigation with Diffusion Trajectory Predictions}
\title{\MYTITLE}

\begin{document}

\begin{spacing}{1.0}
\author{%
Sepehr Samavi, Anthony Lem, Fumiaki Sato, Sirui Chen, Qiao Gu, Keijiro Yano, Angela P. Schoellig, Florian Shkurti
\thanks{Manuscript received: February 11, 2025; Revised May 9, 2025; Accepted June 13, 2025. This paper was recommended for publication by Editor Aniket Bera upon evaluation of the Associate Editor and Reviewers' comments.}
\thanks{Sepehr Samavi, Anthony Lem, Sirui Chen, Qiao Gu, Angela P. Schoellig, and Florian Shkurti are with the University of Toronto Robotics Institute and the Vector Institute for Artificial Intelligence, Toronto, Canada. Angela P. Schoellig is also with the Technical University of Munich and the Munich Institute for Robotics and Machine Intelligence (MIRMI), Munich, Germany. Fumiaki Sato and Keijiro Yano are with Konica Minolta Inc., Japan. (\emph{Corresponding author: Sepehr Samavi} sepehr@robotics.utias.utoronto.ca)}
\thanks{Digital Object Identifier (DOI): see top of this page.}
}
\markboth{IEEE Robotics and Automation Letters. Preprint Version. June, 2025}
{Samavi \MakeLowercase{\textit{et al.}}: SICNav-Diffusion: Safe and Interactive Crowd Navigation with
Diffusion Trajectory Predictions}

\maketitle
\begin{abstract}
To navigate crowds without collisions, robots must interact with humans by forecasting their future motion and reacting accordingly. While learning-based prediction models have shown success in generating likely human trajectory predictions, integrating these stochastic models into a robot controller presents several challenges. The controller needs to account for interactive coupling between planned robot motion and human predictions while ensuring both predictions and robot actions are safe (i.e. collision-free). To address these challenges, we present a receding horizon crowd navigation method for single-robot multi-human environments. We first propose a diffusion model to generate joint trajectory predictions for all humans in the scene. We then incorporate these multi-modal predictions into a SICNav Bilevel MPC problem that simultaneously solves for a robot plan (upper-level) and acts as a safety filter to refine the predictions for non-collision (lower-level). Combining planning and prediction refinement into one bilevel problem ensures that the robot plan and human predictions are coupled. We validate the open-loop trajectory prediction performance of our diffusion model on the commonly used ETH/UCY benchmark and evaluate the closed-loop performance of our robot navigation method in simulation and extensive real-robot experiments demonstrating safe, efficient, and reactive robot motion. Code: \href{https://github.com/sepsamavi/safe-interactive-crowdnav.git}{\footnotesize \texttt{github.com/sepsamavi/safe-interactive-crowdnav.git}}
\end{abstract}

\vspace{-0.5cm}
\section{Introduction} \label{sec:intro}
\begin{figure}[t]
  \footnotesize
  \centering
  \includegraphics[width=0.8\linewidth]{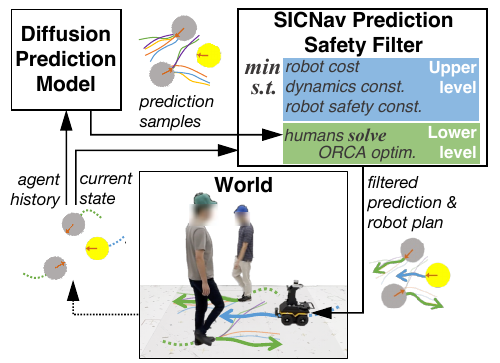}
  \caption{
  For interactive robotic crowd navigation, we use the diffusion prediction model to forecast the future motion of humans as joint trajectory samples. The bilevel SICNav problem uses a particle-filter inspired approach to jointly refine predictions (lower-level) and optimize robot actions that satisfy safety constraints (upper-level). The robot uses solutions in receding horizon fashion. Video: \texttt{\href{http://tiny.cc/sicnav_diffusion}{tiny.cc/sicnav\_diffusion}}}\label{fig:block_diagram}
  \vspace{-0.5cm}
\end{figure}

\IEEEPARstart{F}{or} humans, walking within crowds is considered a trivial task. For robots, however, there exist considerable challenges in achieving safe, collision-free, and efficient crowd navigation.
In an environment like the one illustrated in Fig.~\ref{fig:block_diagram}, a robot must anticipate human movement and adjust its trajectory in real-time. %
This problem has inspired the development of a variety of trajectory prediction methods (e.g. \cite{saltzmann2021trajpp,yuan2021agent,gu2022mid}) that learn distributions of future human trajectories from historical data and demonstrate promising performance in \emph{open-loop} prediction, generating trajectory forecasts for a test set of historical data excluded during training.
However, incorporating open-loop predictions into a robot controller remains a challenge.
Firstly, the robot's actions influence the future motion of humans, and vice versa, i.e. robot and human futures are coupled due to \emph{interaction}. Secondly, the robot needs to account for potentially \emph{multimodal} prediction distributions while choosing only one action to take. Finally, the robot needs to balance \emph{opposing objectives} of efficiently reaching its goal, while ensuring safety.

{\color{blue}
To address interaction challenges, one method first predicts human and robot futures together, then during planning for robot actions penalizes robot deviation from the prediction \cite{poddar2023fromcrowdmotiontorobot}. Yet the plan inevitably deviates as it is optimized towards the robot objective, reducing human prediction relevance.
Other strategies dynamically model predictions, e.g. by representation as affine dynamical systems \cite{Ivanovic2020mats}, or evaluating model gradients during planning \cite{nishimura2020risk,schaefer2021leveraging}. While these methods address prediction multimodality, they expose the critical trade-off between the safety and efficiency. They simplify the probabilistic nature of the planning problem by minimizing a single cost (e.g., expected cost \cite{poddar2023fromcrowdmotiontorobot}, risk metric \cite{nishimura2020risk}, or worst-case cost \cite{schaefer2021leveraging}). In contrast, introducing non-collision constraints maximizes efficiency while ensuring safety.
}

In this work, we introduce \ac{SICNav-Diffusion}, a novel approach for safe robotic crowd navigation, illustrated in Fig.~\ref{fig:block_diagram}. %
We propose the \ac{JMID} prediction model and combine with an extension of \ac{SICNav} \cite{samavi2023sicnav} to simultaneously plan robot actions while filtering the \ac{JMID} predictions for safety. {\color{blue}SICNav embeds \ac{ORCA} \cite{vandenBerg2011orca}, which models human collision-avoidance behavior, as constraints within the robot's planner to form a bilevel \ac{MPC} problem. We extend the bilevel optimization problem such that the \emph{lower-level} \ac{ORCA} refines predictions to ensure non-collision, while the bilevel structure of the \ac{MPC} couples the robot plan with these refined predictions. This way, the optimizer explores trajectory prediction modes generated by \ac{JMID} and selects the one that best optimizes the robot objective, while guaranteeing non-collision.
}

Our contributions are threefold. \underline{First}, we introduce \ac{JMID}, where we extend a single-agent prediction model \acs{MID} \cite{gu2022mid} to model a joint trajectory distribution for multiple humans. \underline{Second}, we propose an extension to \ac{SICNav} \cite{samavi2023sicnav} to use multimodal prediction samples from \ac{JMID}.
\underline{Third}, to evaluate our proposal, we first validate the open-loop prediction performance of \ac{JMID} on the ETH/UCY trajectory benchmark \cite{lerner2007ucydata,pellegrini2009ethdata} and demonstrate \ac{JMID} outperforms the joint prediction performance of state-of-the-art models. We then evaluate the closed-loop planning performance of our method in simulation and extensive {\color{blue} in-lab robot} experiments (240 runs) to compare our approach, SICNav-JMID, with other state-of-the-art prediction models in a single-robot multi-human environment. {\color{blue} Finally, we demonstrate the effectiveness of our method in more than 8 km of outdoor real-world operation.}
\section{Related Work} \label{sec:related_work}
\emph{Trajectory forecasting:} Human trajectory forecasting models use the history of agent positions in a scene to predict their future 2D bird's-eye-view trajectory for some horizon into the future. Recent methods focus on capturing the multi-modality of agent futures using a variety of learning-based architectures, e.g. \acp{CVAE} {\color{blue} with \acp{GRU}} \cite{saltzmann2021trajpp}, autoregressive Transformers \cite{yuan2021agent}, or Diffusion \cite{gu2022mid}. %
Most predict the future of a single human in each inference \cite{saltzmann2021trajpp, gu2022mid}. However, the future of all humans in the scene is coupled, {\color{blue} thus, naively combining the individual forecasts for multiple humans may result in predictions of collisions. }%

{\color{blue}
To address this shortcoming, recent work has focused on jointly predicting all agents in the scene \cite{yuan2021agent, li2024relational, chen2024SPDiff}, for example, by introducing latent variables jointly inferred from all agents \cite{yuan2021agent}, or learning to build relational graphs between groups \cite{li2024relational}. Some methods additionally guide the generative process using social force \cite{Helbing1995sfm} to capture collision-avoidance \cite{yue2022nspsfm, chen2024SPDiff}. While these methods show promising results, guidance does not guarantee collision-free predictions despite adding a bias to the learned model. We extend \cite{gu2022mid} to jointly predict all agents, similar to these methods, however without introducing guidance. Instead, when using the forecasts our robot planner, SICNav, we leverage its interaction model \cite{vandenBerg2011orca} to refine the predictions to be collision-free, with the added benefit of coupling the refined predictions with the robot plan as well.

}

\emph{Combining forecasting with planning:}
{\color{blue}Other works that incorporate predictions into robot planning} focus on distinct challenges: interactivity of planning with predictions, accounting for multimodal prediction distributions, and ensuring safety through non-collision.
{\color{blue} Some methods elect to implicitly model interactions by using \ac{RL} to learn policies that implicitly reason about interactions in a learned policy with explicit \cite{li2024relational,liu2023intention} or implicit \cite{chen_relational_2020,xie2023drlvo} predictions. While these methods can learn rich varied policies, even being deployed outdoors \cite{xie2023drlvo}, it is challenging to evaluate the safety of such black-box policies.

Other} methods \cite{schaefer2021leveraging, Ivanovic2020mats} {\color{blue} explicitly model interactions between agents} by leveraging prediction model gradients \cite{schaefer2021leveraging} in the planning step or outputting predictions as time-varying dynamics used in the planner \cite{Ivanovic2020mats}. While these methods account for multimodal prediction distributions, they incorporate non-collision as an additional objective in the cost function rather than introducing constraints. Other research \cite{lindemann2023safe} ensures probabilistic non-collision constraint satisfaction via conformal prediction but considers unimodal predictions without modeling interactions in the planning step. A blended approach \cite{heuer2023proactive} addresses constraints primarily for the initial MPC step but remains reactive. In contrast, our method addresses the three challenges by selecting prediction modes based on whether the predictions are compatible (i.e. non-colliding) among the humans and with the robot plan, utilizing ORCA to refine the predictions. Our method incorporates this safety-filtering refinement into a bilevel MPC planner to explicitly model interactions between the humans and the robot while guaranteeing that generated action satisfies collision-avoidance constraints with the refined predictions.

\section{Methodology} \label{sec:methodology}
Fig.~\ref{fig:block_diagram} presents a block diagram of our method. In short, %
we first use the history of agent motion to generate a set of \ac{JMID} prediction samples, $\MIDallsamples$. Then, we use the predictions and the latest measured state of the system as the initial state, $\stateinit$, in a Bilevel \ac{MPC} optimization problem that optimizes the robot's actions, $\actionat{0:\horiz-1}$ and system state,
$\stateat{0:\horiz}$, containing the robot states and a refined version of the predicted human trajectories for a finite horizon $\horiz$.
We use this solution in receding horizon fashion: the robot executes the first robot action from the solution, $\actionat{0}$, and repeats the \ac{MPC} optimization with a newly measured state of the system, $\stateinit$, and a newly sampled set of \ac{JMID} predictions, $\MIDallsamples$.

\subsection{SICNav Bilevel MPC Problem} \label{sec:our_soln}
The environment consists of a robot, human agents, indexed by $\idA \in \forallhumanslongset$, and static obstacles in the form of line segments, indexed by $\idStat \in \forallstatobsshortset$.
\label{sec:sysstatedyn}
The state of the system is in continuous space and contains the state of the ego robot, the states of all $\numhumans$ humans, and a set of weights describing the relative importance of each of $\numMIDsamples$ \ac{JMID} samples in determining the intents of the humans.

Formally, at a discrete time step, $t$, the state is denoted as
 	$
    \stateat{t} = (\robstateat{t},    \id{1}{\humstateat{t}}, \dots,  \id{\numhumans}{\humstateat{t}}, \mathbf{w}_t)  \in \sspace,
	$
where the state of the robot,
$\robstateat{t} \in \Reals^{4}$,
consists of its 2D position, heading, and speed. %
The human state,
$\id{\idA}{\humstateat{t}} \in \Reals^{4}$,
consists of 2D position, and 2D velocity $\forallhumans$.
We use a set of importance weights at each time step, $\mathbf{w}_t \in \Reals^\numMIDsamples, \text{ s.t. } \norm{\mathbf{w}_t}_1 = 1$, to calculate a weighted average of $\numMIDsamples$ trajectory prediction samples obtained from \ac{JMID}, which models future trajectories of all humans in the scene, ${\MIDSample} =(\id{1}{\MIDSample},\dots,\id{\numhumans}{\MIDSample}) \in \Reals^{2\horiz\numhumans}$, where $\id{\idA}{\MIDSample} = (\id{\idA}{\MIDsample_{1}}, \dots, \id{\idA}{\MIDsample_{\horiz}})\in\Reals^{2\horiz}$ is the predicted trajectory for individual agent $\idA$, composed of a sequence of predicted 2D positions $\id{\idA}{\MIDsample_{t}}\in\Reals^{2}$. %
We denote the set of samples as $\MIDallsamples = \set{\id{1:\numhumans,s}{\MIDSample}}_{s=1}^{\numMIDsamples}$, and refer to the break down of a particular sample $s$ as $\id{\idA,s}{\MIDSample} = (\id{\idA,s}{\MIDsample_{1}}, \dots, \id{\idA,s}{\MIDsample_{\horiz}})$. We will also refer to the set of all samples for individual agent $\idA$ at a specific time step $t$ as $\id{\idA}{\MIDallsamples_{t}}=\set{\id{\idA,s}{\MIDsample_t}}_{s=1}^{\numMIDsamples}$.

We separate the dynamics of the system into separate functions for the robot, the humans, and the evolution of the importance weights through time. The robot dynamics,
${\robstateat{\kpone}} = \robdyn({\robstateat{t}}, \actionat{t}),$
are modeled as a kinematic unicycle model, where the control input for the system, $\at{\action}{t} \in \cspace$, is a vector of the linear and angular velocity of the robot for time step $t$.
The dynamics of each human $\forallhumans$,
$\id{\idA}{\humstateat{\kpone}} = \humdyn(\id{\idA}{\humstateat{t}}, \id{\idA}{\actionhumat{t}}),$
are modeled as kinematic integrators, where $\id{\idA}{\actionhumat{t}}$ denote actions optimal with respect to \ac{ORCA}.
Note that actions $\id{\idA}{\actionhumat{t}}$ are used to obtain the safety-filtered predictions of the humans. %
Finally, the importance weights evolve according to %
$\mathbf{w}_{\kpone} = \weightsdyn(\stateat{t};\id{1}{\MIDallsamples_{t}},\dots,\id{\numhumans}{\MIDallsamples_{t}})$.
We describe the method for finding the refined predicted human actions and the evolution of importance weights in Section~\ref{sec:diffusion_vpref}.

The stage cost, $\stagecostsymb(\stateat{t}, {\actionat{t}}) = \trans{\stateat{t}} \Q \stateat{t} + \trans{\actionat{t}} \R \actionat{t},$ penalizes deviation from the robot’s goal position and excessive control effort with positive semidefinite matrices $\Q$ and $\R$. The terminal penalty, $\termpenal{\stateat{\horiz}} = \beta \trans{\stateat{\horiz}} \Q \stateat{\horiz},$ where $\beta \geq 1,$ ensures stability of the controller (see Theorem 2.41 in \cite{rawlings2019mpcbook}) with $\beta$ selected to be sufficiently large.

We add constraints of the form $\trans{\at{\state}{t}} \P_\idA \at{\state}{t} \geq \dist_\idA^2, \forallhumans,$ where $\P_\idA \in \Reals^{n \times n}$ extracts the positions of the robot and the $\idA^{th}$ human from the state and $\dist_\idA$ is the minimum distance to avoid collisions. We similarly add quadratic constraints to avoid collisions between the robot and static obstacles, which are represented as piece-wise linear functions. We also bound the input of the system to enforce the kino-dynamic limits of the real-world robot. Finally, for each time-step, we constrain the prediction of each human (indexed $\idA$) to lie in the solution set of the \ac{ORCA} optimization problem \cite{vandenBerg2011orca}, $\id{\idA}{\orcarlxsolnset}(\at{\state}{t};\id{\idA}{\MIDallsamples_{\kpone}})$, described in Section~\ref{sec:orca}. In short, the \ac{ORCA} \ac{QCQP} solves for a predicted human velocity minimizing distance to an intended velocity, obtained from the \ac{JMID} samples for that human, subject to avoiding collisions with other agents and static obstacles.

With these definitions, the \ac{MPC} problem for the robot can be formulated as an optimization problem,
\begin{subequations} \label{eq:combmpc0_prob}
  \allowdisplaybreaks
	\begin{alignat}{2}
		\centremathcell{\minimize_{\substack{\stateat{0:\horiz},\actionat{0:\horiz-1},\\\id{1:\numhumans}{\actionhum_{0:\horiz-1}}, \mathbf{w}_{0:\horiz-1}}}} 	& \centremathcell{\sum_{t=0}^{\horiz-1} \stagecost{\stateat{t},\actionat{t}} + \termpenal{\stateat{\horiz}}} 		& \centremathcell{\quad\quad\quad\quad\quad} \label{eq:combmpc0_min}			\\
		\centremathcell{\subjectto} 																			  							& \centremathcell{\stateat{0}=\stateinit} 																	 		& \centremathcell{\quad\quad\quad\quad\quad} \label{eq:combmpc0_initcondconst}	\\
		\centremathcell{}																										 			& \centremathcell{\robstateat{\kpone}= \robdyn(\robstateat{t},\controlat{t})} 							     		& \centremathcell{\quad\quad\quad\quad\quad} \label{eq:combmpc0_rob_dynconst}	\\ 	%
		\centremathcell{}																										 			& \centremathcell{\id{\idRob}{\action_{min}} \leq {\at{\action}{t}} \leq \id{\idRob}{\action_{\max}}} 	 		& \centremathcell{\quad\quad\quad\quad\quad} \label{eq:combmpc0_actionconst0}	\\ 	%
		\centremathcell{}																										 			& \centremathcell{\Delta\id{\idRob}{\action_{min}} \leq {\at{\action}{t} - \at{\action}{\kmone}} \leq \Delta\id{\idRob}{\action_{\max}}} 	 		& \centremathcell{\quad\quad\quad\quad\quad} \label{eq:combmpc0_actionconst}	\\ 	%
		\centremathcell{}																										 			& \centremathcell{\trans{\at{\state}{t}} \P_\idB \at{\state}{t} \geq {\dist_{\idB}}^2}   	& \centremathcell{\quad\quad\quad\quad\quad} \label{eq:combmpc0_coll_const}	\\ 	%
	  \centremathcell{}																										 			& \centremathcell{\id{\idA}{\actionhumat{t}} \in \id{\idA}{\orcarlxsolnset}(\at{\state}{t};\id{\idA}{\MIDallsamples_{\kpone}})}				 		& \centremathcell{\quad\quad\quad\quad\quad} \label{eq:combmpc0_llorca}		\\ 	%
	  \centremathcell{}																										 			& \centremathcell{\id{\idA}{\humstateat{\kpone}} = \humdyn(\id{\idA}{\humstateat{t}}, \id{\idA}{\actionhumat{t}})}	 		& \centremathcell{\quad\quad\quad\quad\quad} \label{eq:combmpc0_hum_dynconst}	\\	%
    \centremathcell{}																										 			& \centremathcell{\mathbf{w}_{\kpone} = \weightsdyn(\stateat{t};\id{1}{\MIDallsamples_{t}},\dots,\id{\numhumans}{\MIDallsamples_{t}})}	 		& \centremathcell{\quad\quad\quad\quad\quad} \label{eq:combmpc0_weights_dynconst}		%
	\end{alignat}
\end{subequations}
where all the constraints \eqref{eq:combmpc0_rob_dynconst}-\eqref{eq:combmpc0_hum_dynconst} are defined for each time step, $\forallactidcs$, and the constraints \eqref{eq:combmpc0_llorca}-\eqref{eq:combmpc0_hum_dynconst} are defined for each human, $\forallhumans$. The robot collision constraint \eqref{eq:combmpc0_coll_const} is defined for each human agent and each static obstacle in the environment, $\forall \idB \in \set{1,\dots,\numhumans,\dots,\numhumans+\numstatobs}$. %
The constraints \eqref{eq:combmpc0_llorca} are in the solution sets of constrained \ac{ORCA} optimization problems, thus constituting the \emph{lower-level} problems of the bilevel problem \eqref{eq:combmpc0_prob}.

\subsection{Diffusion Model of Joint Motion of Humans} \label{sec:diffusion_intentpred}
\begin{figure}[t]
  \footnotesize
  \centering
  \includegraphics[width=\linewidth]{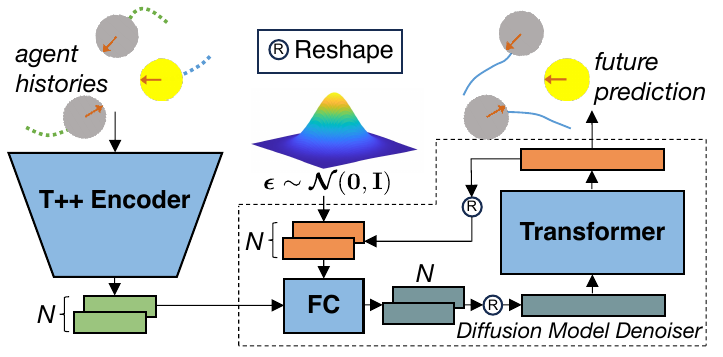}
  \caption{
    Block diagram of \ac{JMID}. Embeddings of agent histories (gray pedestrians and yellow robot) are generated for each agent in parallel by the encoder of \ac{T++}. The embeddings, along with one sample per agent from a multivariate standard normal distribution, are passed in parallel through fully connected (FC) layers. Outputs are concatenated and passed through a Transformer. Denoising (dashed box) is repeated for K denoising steps to generate the final predictions.}
  \label{fig:jmid}
  \vspace{-0.8cm}
\end{figure}

We propose a human trajectory forecasting model that predicts future trajectories for multiple agents jointly. The model generates trajectory samples, $\MIDallsamples,$ which are used to parameterize the the lower-level \ac{ORCA} problems, $\orcarlxsolnset$ in \eqref{eq:combmpc0_prob}.
Our model extends \acs{MID} \cite{gu2022mid}, a diffusion-based trajectory forecaster that predicts future motions for each human individually. The original model is an encoder-decoder network with two components: a borrowed encoder from Trajectron++ (T++) \cite{saltzmann2021trajpp} and a Transformer-based decoder that learns a reverse diffusion process \cite{song2020denoising}. To make a prediction, \acs{MID} requires the history of the target human and its neighboring agents. For the rest of the paper, we refer to \acs{MID} as \ac{iMID}.

We extend iMID to Joint-MID (\ac{JMID}), which produces \emph{joint} samples for \emph{all} humans in the scene,
i.e. each sample contains a forecast of horizon $\horiz$ for all humans.
While \acs{iMID} models the future trajectory distribution of each agent independently, $\id{\idA}{\MIDSample} \sim p(\id{\idA}{\MIDSample}|\stateat{\midhisthoriz:0}),  \forall \idA \in \{1,\dots,\numhumans\}$, JMID models the joint future trajectory distribution of all agents ${\MIDSample} \sim p(\id{1}{\MIDSample}, \dots, \id{\numhumans}{\MIDSample}|\stateat{\midhisthoriz:0})$. A block diagram of \acs{JMID} is illustrated in Fig.~\ref{fig:jmid}.
To obtain joint samples, we first query the encoder in parallel to generate a latent embedding for each human in the scene. We pass these together with noisy future trajectories through fully-connected layers to generate new latent embeddings. For \acs{iMID}, the new embeddings would have been passed through the transformer decoder in parallel. Instead, we concatenate them and then pass them to the transformer. We do not use positional encoding between embeddings so transformer outputs are equivariant to concatenation ordering. Finally, we split the denoised output of the transformer into one prediction per agent, generating the joint samples $\MIDallsamples$. To train the model, we follow \cite{gu2022mid} to maximize the variational lower bound. However, we alter the loss inputs’  dimensions to account for the model output’s varying dimension (i.e. number of humans being predicted) for each scene.

\subsection{Refining Prediction Samples to Satisfy Constraints}
\begin{figure}[t]
  \footnotesize
  \centering
  \includegraphics[width=0.99\linewidth]{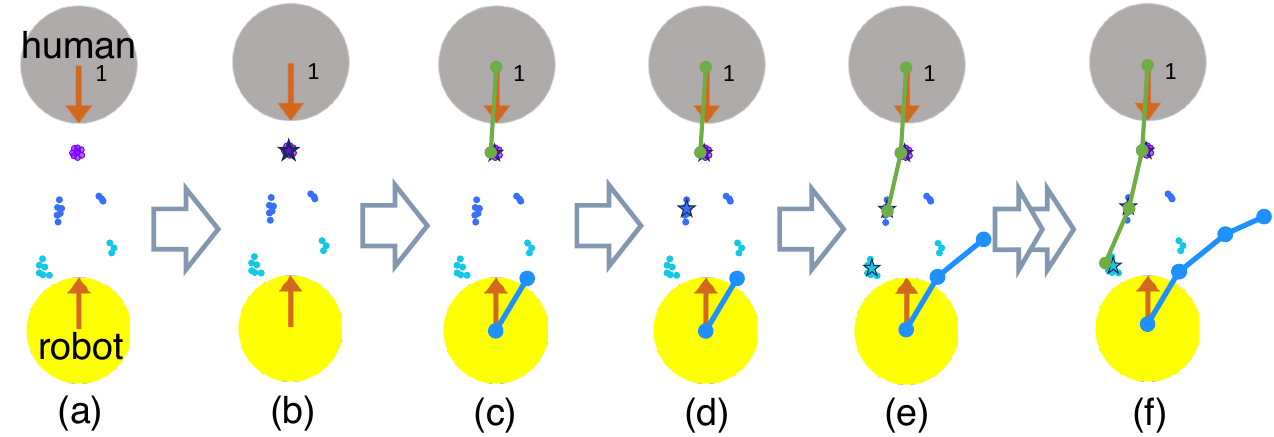}
  \caption{Example walkthrough of steps occurring implicitly within the SICNav-JMID optimization \eqref{eq:combmpc0_prob} demonstrating our method's ability to select a mode of predictions to satisfy safety constraints. (a) Multicolored scatter illustrates \ac{JMID} predictions with $\numMIDsamples=9$ future trajectory samples for human 1 ($\id{1,s}{\MIDsample_1}$ purple to $\id{1,s}{\MIDsample_3}$ cyan). (b) Purple star illustrates the weighted average for the intent $\id{1}{\at{\estpos}{1}}$ found using \eqref{eq:weighted_avg_predpos}. (c) Solution from $t=0$ to $t=1$ with robot plan (blue line) and refined human prediction (green line) results in the robot going right and human left. (d)-(f) After the weights are updated using the human position $\id{1}{\humstateat{1}}$ in \eqref{eq:weight_update_unnormed}-\eqref{eq:weight_update_normed}, the process from (b) and (c) is repeated for subsequent steps resulting in the mode with the human going left being selected in the refined predictions.}
  \label{fig:sample_weighing_example}
  \vspace{-0.7cm}
\end{figure}

\emph{Lower Level ORCA Optimization Problem}: \label{sec:orca}
Within the bilevel problem \eqref{eq:combmpc0_prob}, our method refines the \ac{JMID} predictions to satisfy explicit constraints using the sequences of \ac{ORCA} optimization problems \cite{vandenBerg2011orca} embedded in \eqref{eq:combmpc0_llorca}. %
The \ac{ORCA} optimization problem for agent $\idA$ is formulated as,
\begin{equation} \label{eq:relaxed_orca_argmin}
    \id{\idA}{\orcarlxsolnset}(\at{\state}{t};{\id{\idA}{\at{\MIDallsamples}{\kpone}}}) := \argmin_{{\vel}} \\ \left\{
    \begin{alignedat}{0}
    \centremathcell{\twonorm{{\vel} - {{\vel_{\pref}}}(\stateat{t};{\id{\idA}{\at{\MIDallsamples}{\kpone}}})}^2 \text{}:\text{}}\\
    \centremathcell{\text{collision-avoidance}}
    \end{alignedat}\right\}
\end{equation}
where the optimization variable $\vel \in \Reals^2$ is the velocity of agent $j$ for one time step.
The non-collision constraints are derived using a velocity obstacle approach (see Sec. 4 of \cite{vandenBerg2011orca}) and define linear collision avoidance constraints between agents $\idA$ and agent $\idB\neq\idA$ and for each static obstacle. The constraints are linear with respect to $\vel$ and parameterized by $\stateat{t}$.
The cost is based on a velocity vector from the agent's current position towards the agent's intended position at the next time step, ${\at{\estpos}{t+1}}$, obtained from \ac{JMID} samples,
\begin{equation}
  \vel_{\pref}(\stateat{t}; \id{\idA}{\at{\MIDallsamples}{\kpone}}) = \frac{1}{\delta t} \left(\id{\idA}{\at{\estpos}{t+1}} - \begin{bmatrix}
                                                                                                                          1 & 0 & 0 & 0\\
                                                                                                                          0 & 1 & 0 & 0
                                                                                                                          \end{bmatrix}\id{\idA}{\at{\humstate}{t}}\right),
\end{equation}
where $\delta t$ is the discretization period in \eqref{eq:combmpc0_prob}.
To estimate this intended position, we find the weighted average of the human's predicted positions at time $t+1$ from the \ac{JMID} samples, %
\begin{equation} \label{eq:weighted_avg_predpos}
    \id{\idA}{\at{\estpos}{t+1}} = \sum_{\MIDsampleidx=1}^{\numMIDsamples} \id{s}{\at{\weight}{t}} \id{\idA,\MIDsampleidx}{\at{\predpos}{t+1}}.
\end{equation}
Note that there is one importance weight per joint sample and not per human.
In \eqref{eq:combmpc0_hum_dynconst}, we use the solution of \eqref{eq:relaxed_orca_argmin}, $\id{\idA}{\actionhumat{t}}$, to find the refined predicted position of the agent at the next time step, $\id{\idA}{\at{\humstate}{\kpone}}$.
This way, if the intended velocity for the agent is feasible with respect to constraints in \eqref{eq:relaxed_orca_argmin}, then the optimal velocity found by solving the problem is the one that moves the agent to the estimated intended position at the next time step.
Otherwise, the \ac{ORCA} optimization problem will find the \emph{closest feasible} position to the estimated intent, acting as a safety filter for the \ac{JMID} predictions.
\setlength{\tabcolsep}{3pt}
\begin{table*}
  \caption{Human trajectory forecasting results on the ETH/UCY Dataset. SAD is SADE and SFD is SFDE. All metrics are reported in meters and lower is better $\downarrow$. \textbf{Bold} is best, \nextBest{underline} second-best. {\color{blue} On average \ac{JMID} outperforms the other methds on scene-level metrics.}}
  \label{table:traj_forecast_eth_results}
  \centering
  \footnotesize
  \begin{tabular}{c||cc|cc|cc|cc|cc|cc}
    \toprule
    \multicolumn{1}{c||}{} & \multicolumn{2}{c|}{ETH} & \multicolumn{2}{c|}{Hotel} &
    \multicolumn{2}{c|}{Univ} &
    \multicolumn{2}{c|}{Zara1} &
    \multicolumn{2}{c|}{Zara2} &
    \multicolumn{2}{c}{Average} \\
    \cline{2-13}
    Method & {\scriptsize ADE/FDE} & {\scriptsize SAD/SFD} & {\scriptsize ADE/FDE} & {\scriptsize SAD/SFD} & {\scriptsize ADE/FDE} & {\scriptsize SAD/SFD} & {\scriptsize ADE/FDE} & {\scriptsize SAD/SFD} & {\scriptsize ADE/FDE} & {\scriptsize SAD/SFD} & {\scriptsize ADE/FDE} & {\scriptsize SAD/SFD}\\
    \cline{1-13}
    \ac{T++} & \nextBest{0.53}/0.92 & 0.59/1.09 & \nextBest{0.15/0.23} & \textbf{0.20}/\nextBest{0.35} & \nextBest{0.28}/0.56 & 0.52/1.10 & 0.29/0.57 & 0.38/0.78 & \nextBest{0.16/0.31} & 0.27/0.57 & \nextBest{0.28}/0.52 & 0.39/0.78 \\
    AgentFormer  & \textbf{0.45}/\textbf{0.75} & \textbf{0.48}/\textbf{0.79} & \textbf{0.14}/\textbf{0.22} & \nextBest{0.24}/0.46 & \textbf{0.25}/\textbf{0.45} & 0.62/1.31 & \textbf{0.18}/\textbf{0.30} & \textbf{0.28}/\textbf{0.56} & \textbf{0.14}/\textbf{0.24} & 0.30/0.62 & \textbf{0.23}/\textbf{0.39} & 0.38/0.75 \\
    iMID (DDIM) & 0.63/1.02 & 0.66/1.13 & 0.24/0.41 & 0.28/0.52 & 0.29/0.56 & \nextBest{0.47/1.00} & 0.26/0.52 & 0.35/0.74 & 0.20/0.37 & 0.32/0.66 & 0.32/0.58 & 0.42/0.81 \\
    iMID (DDPM) & 0.54/\nextBest{0.86} & 0.58/0.96 & 0.21/0.35 & 0.26/0.47 & 0.29/\nextBest{0.54} & 0.52/1.10 & 0.24/0.43 & 0.35/0.70 & 0.19/0.33 & 0.33/0.67 & 0.29/\nextBest{0.50} & 0.41/0.78 \\
    \midrule
    JMID (DDIM) & 0.59/1.00 & 0.61/1.05 & 0.21/0.34 & \nextBest{0.24}/0.40 & 0.34/0.70 & \textbf{0.45}/\textbf{0.96} & 0.23/0.45 & \nextBest{0.29}/0.59 & 0.19/0.39 & \nextBest{0.24/0.50} & 0.31/0.58 & \nextBest{0.37/0.70} \\
    JMID (DDPM) & \nextBest{0.53}/0.91 & \nextBest{0.54/0.93} & 0.18/0.27 & \textbf{0.20}/\textbf{0.32} & 0.32/0.66 & \nextBest{0.47}/1.01 & \nextBest{0.22/0.41} & 0.30/\nextBest{0.58} & 0.17/0.32 & \textbf{0.23}/\textbf{0.47} & \nextBest{0.28}/0.51 & \textbf{0.35}/\textbf{0.66} \\
    \bottomrule
  \end{tabular}
  \vspace{-0.5cm}
\end{table*}

\emph{Evolution of Importance Weights}: \label{sec:diffusion_vpref}
At $t=0$, we initialize the importance weights based on the likelihood estimates of the \ac{JMID} samples, which we obtain by fitting \acp{KDE} \cite{Thiede2019KDE}. %
For subsequent steps, in \eqref{eq:combmpc0_weights_dynconst}, we compare the \ac{ORCA}-refined positions, $\id{\idA}{\at{\humstate}{t}}$, found by solving \eqref{eq:relaxed_orca_argmin} at the previous time step with the \ac{JMID} sample's predicted positions, $\id{\idA,\MIDsampleidx}{\at{\predpos}{t}}$,
\begin{equation} \label{eq:weight_update_unnormed}
  \id{s}{\at{\tilde{\weight}}{t}} = \exp\left(-\frac{1}{\numhumans\sigma}\sum_{j=1}^{\numhumans}\twonorm{\id{\idA,\MIDsampleidx}{\at{\predpos}{t}} - \begin{bmatrix}
                                                                                                                                                      1 & 0 & 0 & 0\\
                                                                                                                                                      0 & 1 & 0 & 0
                                                                                                                                                      \end{bmatrix}\id{\idA}{\at{\humstate}{t}}}^2\right),
\end{equation}
where $\sigma$ is a hyperparameter. We normalize these weights,
$
  \id{s}{\at{\bar{\weight}}{t}} = {\id{s}{\at{\tilde{\weight}}{t}}}/{\sum_{s=1}^{\numMIDsamples} \id{s}{\at{\tilde{\weight}}{t}}},
$
then incorporate the influence of the importance weights from the previous step, $\id{s}{\at{{\weight}}{\kmone}}$, and normalize once again,
\begin{equation} \label{eq:weight_update_normed}
  \id{s}{\at{\weight}{t}} = \frac{\id{s}{\at{\weight}{t-1}}\id{s}{\at{\bar{\weight}}{t}}}{\sum_{s=1}^{\numMIDsamples} \id{s}{\at{\weight}{t-1}}\id{s}{\at{\bar{\weight}}{t}}}.
\end{equation}
This way, the importance weights filter out prediction and robot plan combinations that would violate safety constraints.

Fig.~\ref{fig:sample_weighing_example} illustrates an example of refined human predictions and associated robot plan.
For agent 1, we see its \ac{JMID} prediction has two modes: right or left. As the robot jointly optimizes its plan and refines the human predictions through the horizon, at $t=0$ the robot plan moves right. Following the mode where the human also going to the right would result in a collision with the robot's plan (blue line). Since the \ac{ORCA} problem avoids collisions, the refined human prediction moves closer to the left mode as $t$ increases. %
As the variables in \eqref{eq:combmpc0_min} are optimized, the robot plan and \ac{ORCA} cause the human predictions to move into non-colliding prediction modes.

\section{Open-loop Prediction Experiments} \label{sec:openloop_pred_results}

We validate open-loop performance on the popular ETH/UCY benchmark.
Our evaluation follows the standard leave-one-scene-out cross-validation protocol with the standard history length of 3.2s and forecast length of 4.8s. %

\emph{Models}:
{\color{blue} We compare our proposed joint predition method, \textbf{\ac{JMID}} with the following baselines: \textbf{\acs{iMID}} \cite{gu2022mid}, the original individual prediction method as an ablation study on whether \ac{JMID} results in better scene-level performance; \textbf{AgentFormer} \cite{yuan2021agent}, to compare our method with a state-of-the-art autoregressive transformer-based method for joint human trajectory forecasting; and \textbf{\ac{T++}} \cite{saltzmann2021trajpp}, a \ac{CVAE}-based trajectory prediction model, whose encoder is adopted by \acs{iMID} and \acs{JMID}.}
For inference with the diffusion models, \acs{iMID} and \ac{JMID}, we use a \ac{DDPM} \cite{ho2020denoising} version of the models, where denoising is conducted for 100 steps. However, the inference process may not run in real time with \ac{DDPM}s, thus we also evaluate the models with a widely-used accelerated sampling version of the models, \ac{DDIM} \cite{song2020denoising}, where denoising can be conducted with a variable number of steps, 2 in our case. We also use \ac{DDIM} for real-time deployment.

\emph{Metrics}: We evaluate predictive performance using,
\begin{itemize}
    \item Best-of-20 \ac{ADE}: the lowest Euclidean distance between the ground truth trajectory and 20 samples obtained from the model, for each agent.
    \item Best-of-20 \ac{FDE}: the lowest Euclidean distance between positions at the final prediction step, $\horiz$, of the ground truth trajectory and 20 samples obtained from the model, for each agent.
    \item Best-of-20 \ac{SADE} \cite{weng2023joint}: the lowest average \ac{ADE} across all agents in the scene for 20 joint-prediction samples.
    \item Best-of-20 \ac{SFDE} \cite{weng2023joint}: the lowest \ac{FDE} across all agents in the scene for 20 joint-prediction samples.
\end{itemize}
While \ac{ADE} and \ac{FDE} evaluate the performance of the model at the individual agent level (i.e. marginal distributions of each agent's future), \ac{SADE} and \ac{SFDE} evaluate the performance of the model at the scene level (i.e. joint distribution of all agents' futures). \ac{SADE} and \ac{SFDE} only evaluate predictions for different agents that come from the same joint sample, whereas \ac{ADE} and \ac{FDE} can mix-and-match predictions across different joint samples. Our end-goal is to operate a robot among a crowd of humans, so we place a greater emphasis on the scene-level metrics because they evaluate the joint behavior and interactions between agents that a model has learned.

\emph{Implementation}: %
Our \ac{JMID} model uses the same number of layers and dimensions as \ac{iMID} \cite{gu2022mid}.
We train our model using the Adam optimizer with a learning rate of $10^{-4}$ and a batch size of 64. %

\emph{Results and Discussion}:
Table \ref{table:traj_forecast_eth_results} summarizes the results of our experiments. Our method, \ac{JMID}, outperforms the baselines in the scene-level metrics on most test splits and on average. While AgentFormer performs the best in the individual-level metrics, as discussed previously, the scene-level metrics are more meaningful than the individual-level metrics because they evaluate how well a model has learned the interactions between pedestrians. Therefore, our proposed method performs the best in the most meaningful metrics.

\section{Closed-loop Experiments}
\subsection{Simulation Experiments}
\label{sec:sim_exp}
\emph{Testing environment}: We evaluate the performance of our planner in the CrowdSimPlus simulator, a commonly used crowd navigation simulation environment \cite{samavi2023sicnav,Chen2019c}. We generate 500 random initial and final positions for the humans in the environments with a total of $\numhumans=3$ simulated humans. %
The simulated testing area is a $1.75m$-wide corridor with initial positions and goals on either end. In order to ensure a high density of agents in the corridor, we constrain the space using static obstacles, formulated as line-segments. The simulated humans are modeled as \ac{ORCA} agents, however with randomly sampled attributes for radius buffer (unobservable amount added to radius), collision avoidance time horizon (see \cite{vandenBerg2011orca}), goal location, and maximum velocity. None of these attributes are observable to the robot.

\emph{Navigation Methods:} %
{\color{blue} We compare four variants of our SICNav Bilevel \ac{MPC} \eqref{eq:combmpc0_prob}, each using a different human trajectory forecasting model: \textbf{SICNav-JMID} using our prosposed \ac{JMID} samples, \textbf{SICNav-iMID} using individual agent samples from \ac{iMID} \cite{gu2022mid}, \textbf{SICNav-AF} using joint samples from AgentFormer \cite{yuan2021agent}, and \textbf{SICNav-CVG} a non-learning-based constant-velocity-goal baseline which projects the human's current velocity forward in time to infer their goal. To evaluate the effect of SICNav's bilevel modeling of interactions we compare with \textbf{MPC-CVMM} which has the same cost and constraint functions as SICNav except using \ac{CVMM} human predictions that remain constant during the \ac{MPC} optimization, neglecting interactions. Nonetheless \ac{CVMM} predictions have been shown to be a competitive paradigm for prediction \cite{schoeller2020cvmm}. We also compare our method to the reinforcement learning methods \acl{SARL} (\textbf{\acs{SARL}}) \cite{Chen2019c} and \acl{RGL} (\textbf{\acs{RGL}}) \cite{chen_relational_2020}, as well as a classical reactive baseline, \acl{DWA} (\textbf{\acs{DWA}}) \cite{Fox1997dwa} which was used on some of the oldest crowd-navigating robots.}

\emph{Implementation}: The prediction models \ac{JMID}, \ac{iMID}, and AgentFormer are implemented in PyTorch and wrapped with ROS to deploy inference on the robot at 10 Hz. We pretrain the models on the JRDB dataset \cite{martin2021jrdb}, which contains a robot agent in its scenes. The training details are the same as Sec.~\ref{sec:openloop_pred_results}. We then fine-tune the models using data collected from training scenarios in the simulator. %

\emph{Performance metrics}: To evaluate robot navigation performance, we adopt the following metrics from literature \cite{samavi2023sicnav}:
\begin{itemize} %
	\item \textit{Success Rate:} The rate of scenarios in which the robot is able to arrive at its goal within $30s$. It measures the success of the robot's navigation through the crowd.
	\item \textit{Average Navigation Time:} The average time to completion, which indicates the robot's navigation efficiency.
	\item \textit{Collision Frequency:} The frequency the robot is in a collision state with another agent, which indicates the safety of the robot.
	\item \textit{Freezing Frequency:} The frequency of the robot stopping completely, which measures how often the robot can potentially get stuck in a deadlock with the human agents (encounters the \ac{FRP}).
\end{itemize}

\emph{Results and Discussion}: Table~\ref{tab:sim_results} summarizes the results of our simulation experiments. SICNav-JMID outperforms the other methods in success rate and freezing frequency. The results suggest the joint prediction of human trajectories with \ac{JMID} is more effective than individual predictions with \ac{MID} or the joint predictions from AgentFormer. The results also suggest the simple projection of human velocity with SICNav-CVG is not as effective as the learning-based methods {\color{blue} resulting in a lower success rate and higher navigation time. The MPC-CVMM baseline has a lower navigation time than the SICNav variants, however at the cost of a much higher collision frequency. SARL and RGL have a much lower success rate and freeze more often than the SICNav. The reactive DWA method has the lowest success rate, highest navigation time, and the highest collision frequency.}
\begin{table}[t]
  \caption{Quantitative comparison in 500 randomly generated test scenarios with 1 robot and 3 simulated humans following a randomized \ac{ORCA} policy. Arrows indicate desired relative values. \textbf{Bold} is best, \nextBest{underline} second-best.}  \label{tab:sim_results}
  \begin{center}
      \footnotesize
      \begin{tabular}{|l|c|c|c|c|} %
          \hline
          \multicolumn{1}{|p{1.2cm}|}{\centering \vfill \textbf{Approach}} & \multicolumn{1}{p{1.0cm}|}{\centering \textbf{Success Rate} $\uparrow$} &  \multicolumn{1}{p{1.3cm}|}{\centering \textbf{Avg Nav Time (s)} $\downarrow$} & \multicolumn{1}{p{1.7cm}|}{\centering \textbf{Collision Freq. (s$^{-1}$)} $\downarrow$} & \multicolumn{1}{p{1.7cm}|}{\centering \textbf{Frozen Freq. (s$^{-1}$)} $\downarrow$} \\
          \hline
          SICNav-JMID & $\mathbf{1.00}$ & $\underline{6.82}$          & $\underline{0.006}$          & $\mathbf{0.20}$\\
          SICNav-iMID & $\underline{0.98}$        & $7.54$          & $0.007$          & $\underline{0.38}$ \\
          SICNav-AF   & $0.97$          & $7.67$          & ${0.007}$        & $0.43$ \\
          SICNav-CVG  & $0.95$          & $8.86$          & $\textbf{0.005}$ & $0.73$ \\
          \color{blue} MPC-CVMM                        & \color{blue} $0.97$ & \color{blue} $\mathbf{6.53}$ & \color{blue}$0.018$ & \color{blue}$0.55$ \\
          \color{blue} SARL \cite{Chen2019c}           & \color{blue} $0.89$ & \color{blue} ${7.09}$ & \color{blue}$\underline{0.006}$ & \color{blue}$0.96$ \\
          \color{blue} RGL \cite{chen_relational_2020} & \color{blue} $0.89$ & \color{blue} ${6.92}$ & \color{blue}${0.007}$ & \color{blue}$1.62$ \\
          \color{blue} DWA \cite{Fox1997dwa}           & \color{blue} $0.85$ & \color{blue} ${11.52}$ & \color{blue}$0.032$ & \color{blue}$0.93$ \\
          \hline
      \end{tabular}
  \end{center}
  \vspace{-0.4cm}
\end{table}
\subsection{{\color{blue}In-Lab} Robot Experiments} \label{sec:robot_experimetns}
\begin{figure}[t]
	\centering
	\footnotesize
	\begin{subfigure}[]{0.49\linewidth}
		\centering
    \footnotesize
		\includegraphics[width=0.95\linewidth]{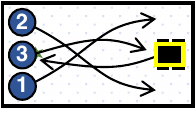}
		\caption{Variant 1}
    \label{fig:real_robot_var1}
	\end{subfigure}
	\hfill
  \begin{subfigure}[]{0.49\linewidth}
		\centering
    \footnotesize
		\includegraphics[width=0.95\linewidth]{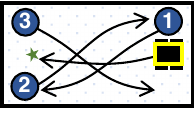}
		\caption{Variant 2}
    \label{fig:real_robot_var2}
	\end{subfigure}
  \caption{Real-robot experiment scenarios with variants 1 (a) and 2 (b). The green star indicates the robot's goal, numbered circles indicate human start positions, and arrows their goals. In a scenario with $\numhumans \in \set{1,2,3}$ humans, positions greater than $\numhumans$ are empty}
  \label{fig:real_robot_scenarios}
  \vspace{-0.6cm}
\end{figure}

\begin{figure}[t]
	\centering
	\footnotesize
	\begin{subfigure}[]{0.49\textwidth}
		\centering
    \footnotesize
		\includegraphics[width=\textwidth]{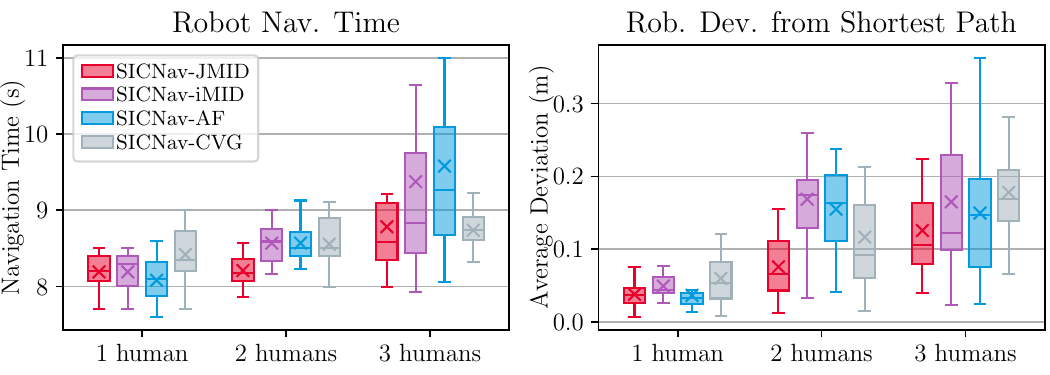}
		\caption{Robot Navigation Efficiency}
    \label{fig:real_robot_nav_perf_results}
	\end{subfigure}
	\hfill
  \begin{subfigure}[]{0.49\textwidth}
  \centering
  \footnotesize
  \includegraphics[width=\textwidth]{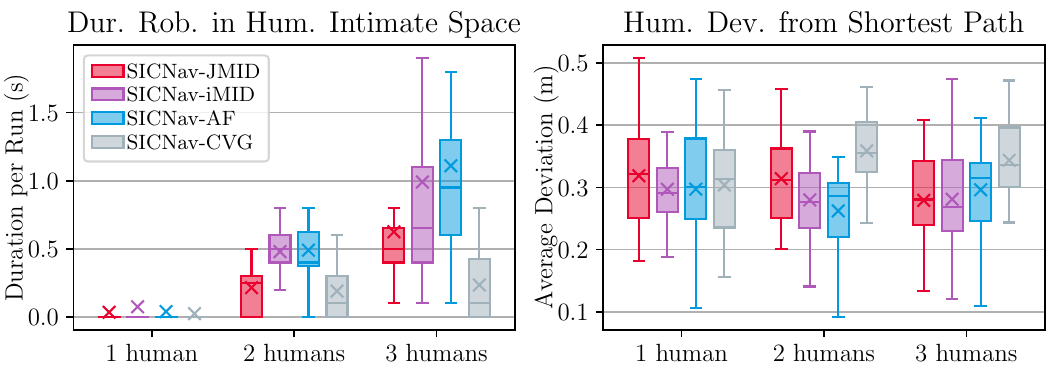}
  \caption{Robot-Human Proxemics}
  \label{fig:real_robot_safety_results}
  \end{subfigure}
  \caption{{Box-and-whisker plots of robot navigation performance in terms of (a) efficiency and (b) proxemics in scenarios with varying numbers of humans. Median values are illustrated by the line in the box-and-whisker plots and mean values are indicated by the $\times$. Our method SICNav-JMID (a) results in the most efficient robot navigation with respect to both metrics, and (b) spends the least time in the intimate space of humans compared to other learning-based methods. SICNav-CVG spends less time in intimate space because it causes humans to deviate from their shortest paths at higher human densities.}}
  \label{fig:real_robot_closedloop_results}
  \vspace{-0.75cm}
\end{figure}

\begin{figure}[th]
	\centering
	\footnotesize
  \begin{subfigure}[]{0.49 \textwidth}
		\centering
	    \footnotesize
		\includegraphics[width=\textwidth]{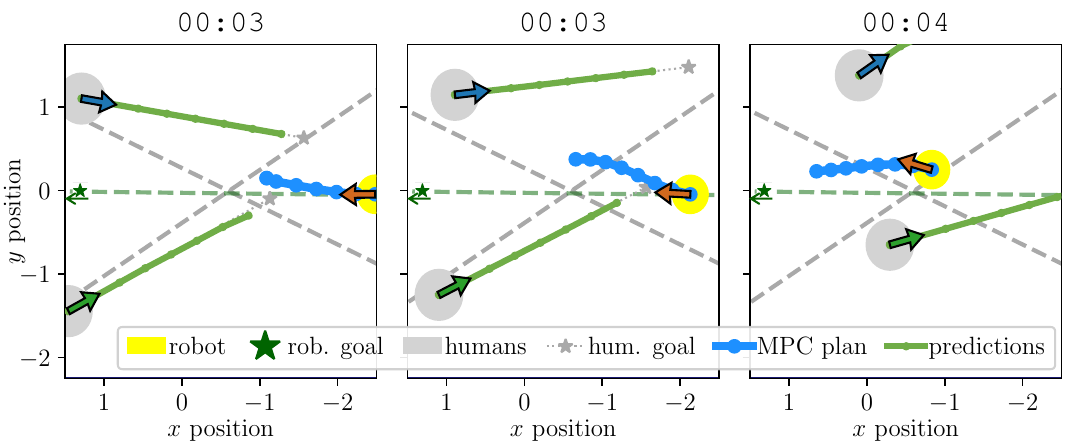}
		\caption{Top view snapshots of SICNav-CVG robot}
	    \label{fig:qual_sicnav_cvg_snapshots}
	\end{subfigure}
  \hfill
  \begin{subfigure}[]{0.49 \textwidth}
		\centering
	    \footnotesize
		\includegraphics[width=\textwidth]{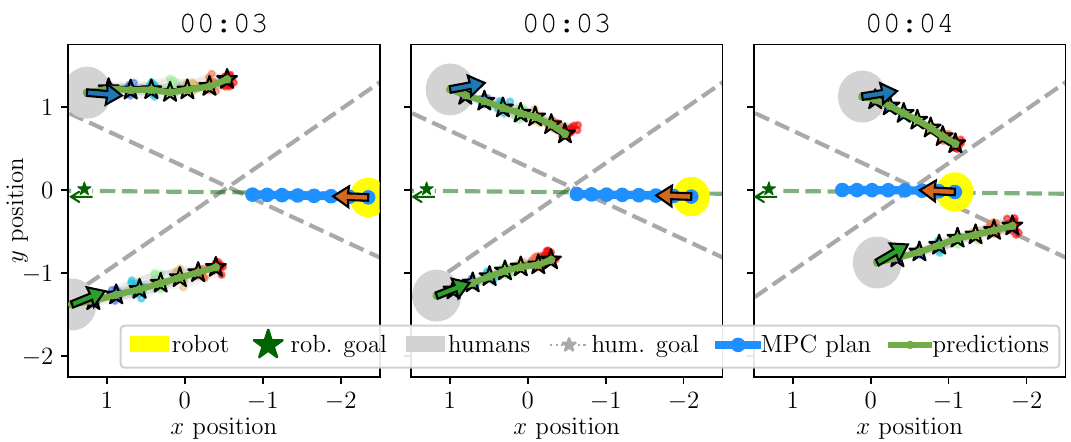}
		\caption{Top view snapshots of SICNav-JMID robot}
	    \label{fig:qual_sicnav_midjp_snapshots}
	\end{subfigure}
    \caption{%
    {\color{blue} We illustrate two similar scenarios with SICNav-CVG (a) and SICNav-JMID (b). We plot temporal snapshots of the scenarios with dashed lines (gray for humans, green for robot) indicating each agent's shortest-path-to-goal. The SICNav-CVG robot produces solutions that drive towards the blue-hat human, causing both the human and robot to deviate from their shortest paths, while the JMID predictions cause SICNav-JMID to produce solutions where all agents stay closer to their shortest-paths-to-goal.}
    } \label{fig:qual_pred}
    \vspace{-0.75cm}
\end{figure}

\emph{Testing Environment:} We also evaluate the performance of our approach on a real robot in a controlled environment similar to \cite{samavi2023sicnav}. We use a Clearpath Jackal differential drive robot pictured in Fig.~\ref{fig:block_diagram}. The robot weighs $20kg$ and measures $46cm$W$\times60cm$L$\times50cm$H. Our operating environment is an indoor space measuring $5m$$\times$$9m$. We conduct a total of 240 runs with a varying number of humans $\numhumans$$\in$$\set{1,2,3}$ and two configuration variants, illustrated in Fig.~\ref{fig:real_robot_scenarios} (i.e. six scenarios).
For each run of a particular scenario, all human agents were instructed to move from their labelled starting positions to the opposite corner of the room (illustrated by the arrows). In variant 1 (Fig.~\ref{fig:real_robot_var1}) all humans start from the opposite side of the room from the robot while in variant 2 one human moves in the same direction as the robot (Fig.~\ref{fig:real_robot_var2}). These scenarios are designed to cause a conflict zone in the middle of the room where all agents need to interact. We repeat 10 runs per variant and navigation method being tested. We use a VICON motion capture system to measure the positions of the robot and human agents in the environment and use an \ac{EKF} and \acp{KF} to track the positions and velocities of the robot and humans, respectively.

\emph{Navigation Methods:} We evaluate the {\color{blue} four variants of SICNav tested} in the simulation experiments: SICNav-JMID, SICNav-iMID, SICNav-AF, and SICNav-CVG.

\emph{Implementation}: We use the \ac{ROS} 1 middleware. The robot accepts linear and angular velocity commands to track by its low level controller. We run the algorithms on a {\color{blue} laptop with 24-core Intel Core i9-14900HX CPUs and an NVIDIA RTX 4080 GPU. The prediction models are fine-tuned as in the simulation experiments, however with data collected with the robot. The inference code for the models is wrapped in a ROS node that publishes the most recent prediction samples at 10 Hz, with inference taking $<0.1s$ using 2-step \ac{DDIM}, as in Sec.~\ref{sec:openloop_pred_results}}. The controller ROS node also runs the controller in real-time replanning at 10 Hz {\color{blue}(solve-times taking $<0.1s$)} with CasADi as the modeling language and the Acados \cite{verschueren2021acados} solver for optimization. %
We discretize time at $\delta t = 0.25s$ and use an \ac{MPC} horizon of $\horiz = 2s$.

\emph{Results and Discussion:} %
We evaluate the robot's navigation efficiency using two metrics: the deviation from the shortest path to the goal and the robot's navigation time (i.e. time to reach the goal in each run). Fig.~\ref{fig:real_robot_nav_perf_results} shows the box-and-whisker plots of the robot navigation efficiency metrics for the different navigation methods. We observe for both metrics, in scenarios with $\numhumans = 1$, all methods behave similarly. However, as the number of humans increases to $\numhumans \in \set{2,3}$, using SICNav-JMID results in both a lower deviation from the shortest path to the goal and a lower navigation time compared to the other methods.

We also analyze the interaction patterns between the robot and humans when using the different navigation methods. First, we analyze the effect of the robot's navigation on the humans by looking at the deviation of the humans from their shortest paths to the goal. Fig.~\ref{fig:real_robot_safety_results} shows the box-and-whisker plots of the humans' deviation from their shortest paths to the goal for the different navigation methods (right). We observe in all three scenarios, the learning-based methods SICNav-JMID, SICNav-iMID, and SICNav-AF perform similarly on this metric. However, in the scenarios with $\numhumans \in \set{2,3}$, SICNav-CVG causes a higher deviation of the humans from their shortest paths to the goal.%

Second, to analyze proxemics, we look at the duration of time the robot spends within the \emph{intimate space} \cite{hall1990proxemics}, i.e. when the robot is within $0.45m$ of a human. Fig.~\ref{fig:real_robot_safety_results} illustrates the box-and-whisker plots of the duration of time the robot spends in the intimate space of a human per run for the different navigation methods (left). In the scnarios with $\numhumans$$=$$1$, we see all methods spend almost no time in the personal space of the human. In the scenarios with $\numhumans$$\in$$\set{2,3}$, we observe SICNav-JMID spends less time in the intimate space of the humans compared to the other learning-based methods, SICNav-iMID and SICNav-AF. However, the non-learning-based baseline SICNav-CVG spends the least time in the personal space of the humans. Recalling the results from Fig.~\ref{fig:real_robot_safety_results}, we see that this lower amount of time spent in the personal space of the humans comes at the cost of a higher deviation of the humans from their shortest paths.%

We interpret this result as meaning SICNav-CVG causes the humans to \emph{run away} or \emph{avoid} getting close to the robot{\color{blue}, which is undesirable}. We illustrate an example occurrence of this phenomenon from the runs of the experiments in Fig.~\ref{fig:qual_pred}.
{\color{blue}
We analyze two similar scenarios with SICNav-CVG (\ref{fig:qual_sicnav_cvg_snapshots}) and SICNav-JMID (\ref{fig:qual_sicnav_midjp_snapshots}). In the first snapshots for both methods, we observe the blue-hat agent's heading (blue arrow) deviates more from their optimal path (gray dashed line) than the green-hat agent's heading. With CVG predictions (\ref{fig:qual_sicnav_cvg_snapshots}), the blue hat agent is predicted to deviate far from their optimal path. As a result, at the next control step (middle snapshot), the SICNav-CVG robot turns toward the blue-hat agent forcing the blue-hat agent to deviate even further from their optimal path in the third control step (final graph in the snapshots). In contrast, with JMID (\ref{fig:qual_sicnav_midjp_snapshots}) predictions, at the second control step (middle graph in the snapshots), the JMID model predicts the blue-hat agent intends to return toward their optimal path, despite the agent's heading pointing away from the path. As such, the robot's MPC plan does not turn towards the blue-hat agent, allowing the robot and blue-hat agent to stay nearly optimal.}

\begin{figure}
  \centering
  \footnotesize
  \includegraphics[width=0.49\textwidth]{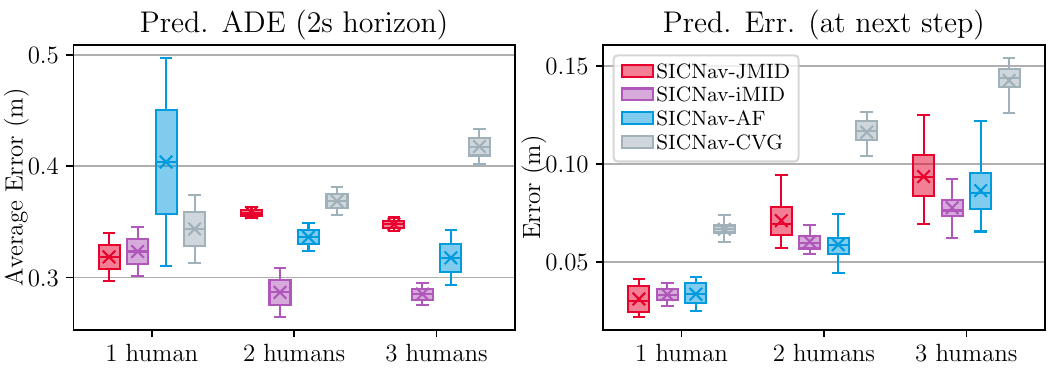}
  \caption{Box-and-whisker plots of prediction performance in the closed loop experiments. We evaluate for all pedestrians \acf{ADE} over the \ac{MPC} horizon (left) and prediction error at the next step the robot is used (right). The learning-based methods outperform the CVG baseline in terms of both metrics. Surprisingly, despite outperforming all methods in terms of navigation performance, the \ac{JMID} model does not consistently outperform the other learning-based methods in terms of the prediction metrics.}
  \vspace{-0.6cm}
  \label{fig:real_pred_perf}
\end{figure}
Finally, we evaluate prediction performance in the closed-loop experiments. Fig~\ref{fig:real_pred_perf} illustrates the \acp{ADE} over the 2s horizon (left) as well as the predictive error at the next control step (right). To describe the latter metric, recall we use the \ac{MPC} solutions in receding horizon fashion, thus the action from a solution obtained at time $t$ is only used until the next solution becomes available at $t+\delta t$. We evaluate the predictive error at this moment as it is the latest time the solution is used by the robot. We see for both metrics, the learning-based methods outperform the CVG baseline.
Surprisingly, although the trajectory prediction results from JMID do not consistently outperform the other learning-based methods in every scenario, JMID nonetheless achieves the best closed-loop navigation performance.
This observation highlights the importance of joint prediction in crowd navigation, yet also suggests the possible mismatch between open-loop prediction metrics and closed-loop robot navigation performance, as also shown in \cite{poddar2023fromcrowdmotiontorobot}.
{\color{blue}
\subsection{Outdoor Robot Experiments} \label{sec:outdoor_robot}
We deploy a robot running SICNav-JMID outdoors using only on-board sensing to demonstrate the viability of SICNav-MID in the absence of a motion capture system for perception and localization and in the face of larger crowds than the in-lab experiments outlined in Sec.~\ref{sec:robot_experimetns}. We use a Clearpath Jackal robot only using two sensors: an Ouster OS1-128 lidar and wheel odometery. For \ac{SLAM} we use Cartographer \cite{hess2016cartograper}. For human perception and tracking, we use YOLOv9 to generate 2D detections on images generated by the lidar \cite{wang2024yolov9} and aUToTrack for human tracking in 3D \cite{burnett2019autotrack}. For obstacle map generation and global path planning we use the \ac{ROS} 1 \texttt{move\_base} framework. We run SICNav-JMID with implementation details and compute as in Sec.~\ref{sec:robot_experimetns}. The computer also runs the object detection and tracking software but we run sensor driver and localization software on the robot's on-board computer.

We operate the robot in sidewalk and campus environments in 21 scenarios for a total 8.86 km over 2 hours 26 minutes. In each scenario, an operator provides a robot goal on the map using a wireless tablet. The robot then generates a global path to the goal and SICNav-JMID begins tracking the path. In our experiments, the robot encountered up to 8 pedestrians at a time and required minimal manual takeovers, successfully driving autonomously 99.03\% of the time. We provide qualitative analysis in the video: \texttt{\href{http://tiny.cc/sicnav_diffusion}{tiny.cc/sicnav\_diffusion}}.
}
\section{Conclusion}
\label{sec:conclusion}
In this work we propose a novel method for joint human trajectory forecasting and robot navigation in crowded environments. We introduce the \ac{JMID} model, which generates joint trajectory forecasts for all agents in the scene. We then propose a Bilevel \ac{MPC} method that uses these forecasts to simultaneously refine the predictions for safety and optimize the robot's actions. We validate the open-loop performance of our prediction model on the ETH/UCY benchmark and show it outperforms state-of-the-art methods in terms of joint forecasting. We also conduct real-robot experiments {\color{blue} to show our method outperforms state-of-the-art methods in terms of robot navigation efficiency and safety} in a controlled environment {\color{blue} and demonstrate the viability of our method in uncontrolled outdoor environments}.

{\footnotesize

\begin{spacing}{0.95}
  \color{blue}
\printbibliography
\end{spacing}
}

\end{spacing}
\end{document}